\documentclass[letterpaper, 10 pt, conference]{ieeeconf}

\IEEEoverridecommandlockouts
\usepackage{cuted}
\usepackage{placeins}
\usepackage{graphicx}
\usepackage{amsmath,amssymb,amsfonts}
\usepackage{algpseudocode}
\usepackage{float}
\usepackage{algorithm}
\usepackage{textcomp}
\usepackage{xcolor}
\usepackage{hyperref}
\usepackage{subcaption}
\usepackage{caption}
\usepackage{booktabs}
\usepackage{comment}
\usepackage{gensymb}
\usepackage{cleveref}
\usepackage{flushend}
\usepackage{pgfplots}
\pgfplotsset{compat=1.17}
\usepackage{stfloats}
\usepackage{multirow}
\fnbelowfloat
\definecolor{EyeInFingerColor}{RGB}{0, 102, 204}
\definecolor{RealsenseColor}{RGB}{204, 51, 0}

\title{\LARGE \bf
CAD-Prompted SAM3:\\ Geometry-Conditioned Instance Segmentation for Industrial Objects
}

\author{
Zhenran Tang,
Rohan Nagabhirava,
Changliu Liu%
\thanks{\scriptsize Zhenran Tang, Rohan Nagabhirava, and Changliu Liu are with the Robotics Institute, Carnegie Mellon University, Pittsburgh, PA, USA ({\tt\footnotesize \{zhenrant,rnagabhi,cliu6\}@andrew.cmu.edu}). In collaboration with Lockheed Martin Advanced Technology Laboratories (LM ATL). Research was sponsored by the ARM (Advanced Robotics for Manufacturing) Institute through a grant from the Office of the Secretary of Defense and was accomplished under Agreement Number W911NF-17-3-0004. The views and conclusions contained in this document are those of the authors and should not be interpreted as representing the official policies, either expressed or implied, of the Office of the Secretary of Defense or the U.S. Government. The U.S. Government is authorized to reproduce and distribute reprints for Government purposes notwithstanding any copyright notation herein.}
}

\begin{document}
\maketitle

\vspace{-5.0em}
\begin{strip}
  \centering
  \vspace{-7.8em} % optional: tighten to title
  \includegraphics[width=\textwidth]{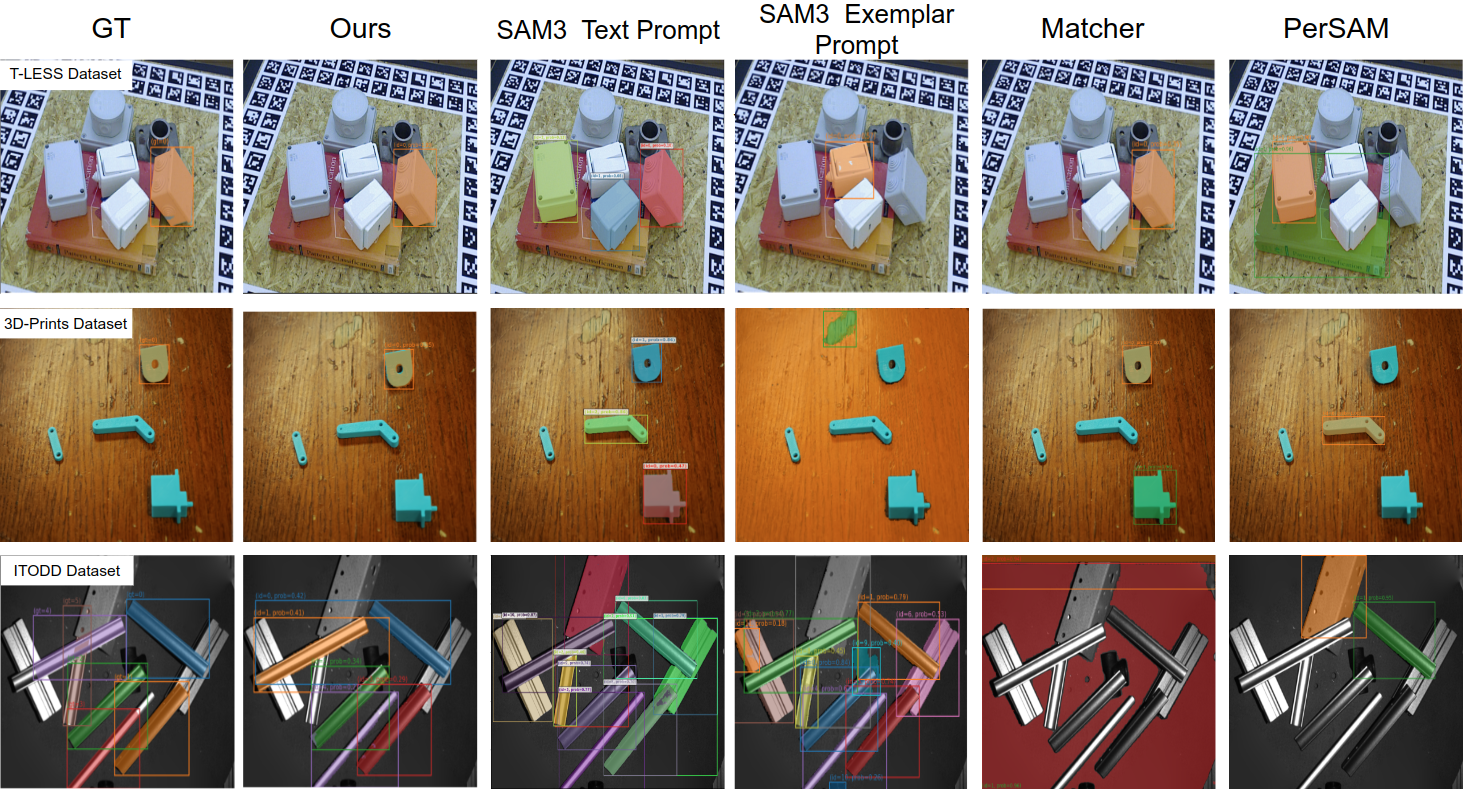}
  \captionof{figure}{Qualitative comparison of CAD-prompted SAM3 with SAM3 (exemplar and text prompts), Matcher, and PerSAM. Our approach produces more accurate instance masks given identical CAD-based prompts. Text prompts for SAM3 are generated from the CAD renderings using GPT-5.1.}
  \label{fig:teaser}
  \vspace{-0.8em} % optional: tighten to abstract
\end{strip}

\begin{abstract}
Language-prompted segmentation is inherently limited by the expressiveness of natural language and struggles with uncommon, instance-specific, or difficult-to-describe objects. These scenarios are frequently encountered in manufacturing and 3D printing environments. While image exemplars provide an alternative, they primarily encode appearance cues such as color and texture, which are often unrelated to a part’s geometric identity. In industrial settings, a single component may be produced in different materials, finishes, or colors, making appearance-based prompting unreliable. In contrast, such objects are typically defined by precise CAD models that capture their canonical geometry. 
We propose a CAD-prompted segmentation framework built on SAM3 that uses canonical multi-view renderings of a CAD model as prompt input. The rendered views provide geometry-based conditioning independent of surface appearance. The model is trained using synthetic data generated from mesh renderings in simulation under randomized appearances. Our approach enables single-stage, CAD-prompted mask prediction, extending promptable segmentation to objects that cannot be robustly described by language or appearance alone.

\end{abstract}
{\small
\noindent
Code, sample dataset, and checkpoint are available at\\
\url{https://github.com/kevinPOI/cad_prompted_sam3}
}
\section{Introduction}
Robotic manipulation in cluttered environments depends on reliable perception to identify target objects before planning interaction. Instance segmentation is commonly used to separate objects from background and neighboring items, producing masks that can be used to filter point clouds for pose estimation and grasp selection. Integrated systems in robotic competitions such as RoboCup@Home~\cite{robocup} and the Amazon Picking Challenge~\cite{cartman} include segmentation as part of the perception pipeline for object localization and manipulation. In practice, YOLO-style models ~\cite{yolo} provide accurate and efficient detection and segmentation suitable for real-time robotic applications. However, achieving reliable performance on new object categories typically requires collecting and annotating hundreds of instances per class. In rapid prototyping and small-batch manufacturing settings, where new parts are introduced frequently and produced in small quantities, this data requirement can be a significant hurdle.

Recent advances in promptable segmentation, particularly foundation models such as the Segment Anything Model~3 (SAM3)~\cite{sam3}, demonstrate strong generalization across open-world imagery and support both language and image-exemplar prompting. Language-prompted vision models perform well for broad semantic categories but are less suitable for engineered components, which often lack descriptive names and are distinguished by subtle geometric differences that vision language models have difficulty resolving~\cite{peng2024finegrained, ju2025evaluation}.

Image-exemplar prompting reduces linguistic ambiguity by conditioning on visual references. However, exemplar-based methods degrade when substantial intra-class appearance variation exists between support and query images~\cite{chen2024learning}. In rapid prototyping and small-batch manufacturing, this is a significant limitation, as the same component may be produced in different colors, materials, or surface finishes.

In contrast, engineered objects are canonically defined by their CAD models. A CAD file specifies the exact geometry of a component independent of its surface appearance. In this work, we introduce a CAD-prompted segmentation framework that extends promptable segmentation to geometry-defined objects. Building upon SAM3, we use canonical multi-view renderings of a CAD model as prompt inputs to condition segmentation on geometric structure while remaining independent of surface appearance. The model is trained using large-scale synthetic data generated by rendering mesh-based objects with randomized color and textures, to enable instance segmentation for objects that are difficult to describe by language or exemplar images, as shown in Fig.~\ref{fig:teaser}. Our contributions are:

\begin{itemize}
    \item We formulate CAD-prompted promptable segmentation, introducing canonical multi-view renderings as a geometry-aware prompt modality.
    \item We present a synthetic training pipeline that leverages mesh renderings to disentangle geometry from appearance variation.
    \item We demonstrate that CAD prompts enable robust single-stage dense mask prediction for industrial objects across varying materials and surface properties, and support zero-shot robotic manipulation of unseen objects.
\end{itemize}

By conditioning segmentation on CAD-defined geometry rather than semantic labels or exemplar appearance, our framework aligns perception with how engineered components are specified in manufacturing practice. This enables reliable open-set instance segmentation for parts whose identity is determined by shape rather than color, texture, or language.

\section{Related Work}

\subsection{Foundation Segmentation Models}

Foundation segmentation models replace fixed semantic classifiers with flexible input prompts such as points, boxes, language, or image exemplars. 
The original Segment Anything Model (SAM) introduced large-scale prompt-conditioned mask prediction and demonstrated strong generalization across datasets~\cite{sam1}.  
SAM3 further extends this framework to promptable concept segmentation, enabling mask prediction conditioned on text and image exemplars across images and videos~\cite{sam3}. 
These works show that segmentation can be formulated as a general prompt-conditioned task. 
However, when conditioned on visual exemplars, the representation remains based on appearance similarity.

\subsection{Appearance-Based Exemplar Conditioning}

Several works perform cross-image segmentation by conditioning on a visual exemplar. 
Earlier few-shot segmentation methods such as CANet and PFENet compute dense feature correspondences between support and query images, using the resulting similarity maps to guide subsequent mask prediction modules~\cite{canet, tian2020pfenet}. 
Within the SAM framework, PerSAM computes similarity in SAM’s image embedding space to derive explicit point prompts, along with high-level semantic prompts and target-guided attention to improve mask prediction~\cite{zhang2023persam}. 
Matcher performs bi-directional feature matching using an external DINOv2~\cite{dinov2} backbone to generate point prompts, and then applies SAM for mask prediction~\cite{matcher}. 
Despite architectural differences, these methods rely primarily on explicit dense support--query feature similarity to guide segmentation.

\subsection{CAD-Based Detection with Proposal Matching}

Modern model-based, unseen object detection methods on RGB images typically generate object proposals and match them against rendered views of a predefined set of CAD models using pretrained visual features.
Representative systems include CNOS, which first uses SAM~\cite{sam1} to segment all possible object instances as proposals, then scores proposals against CAD-rendered views using DINO~\cite{dinov1} features to identify object instances~\cite{cnos}. 
Subsequent works such as MUSE~\cite{muse} and NIDS-Net~\cite{nids-net} improve this pipeline by adopting stronger proposal models and feature backbones, and refining similarity scoring.
These approaches evaluate multiple object candidates per image and perform per-object scoring against a set of CAD models, rather than providing a unified promptable segmentation interface.

\textbf{Take Away: }
Existing exemplar-based methods rely on support–query similarity maps to guide segmentation, while proposal-matching pipelines use CAD models only after mask generation, scoring object proposals against a predefined CAD model set within a two-stage detection framework.
In contrast, we formulate CAD-prompted perception as a single-stage promptable segmentation problem. 
Built on SAM3~\cite{sam3}, our method replaces RGB exemplars with canonical multi-view renderings of CAD models to allow geometry-conditioned mask predictions.
\section{Methodology}

\begin{figure*}
    \vspace{0.05in}
    \centering
    \includegraphics[width=0.98\linewidth]{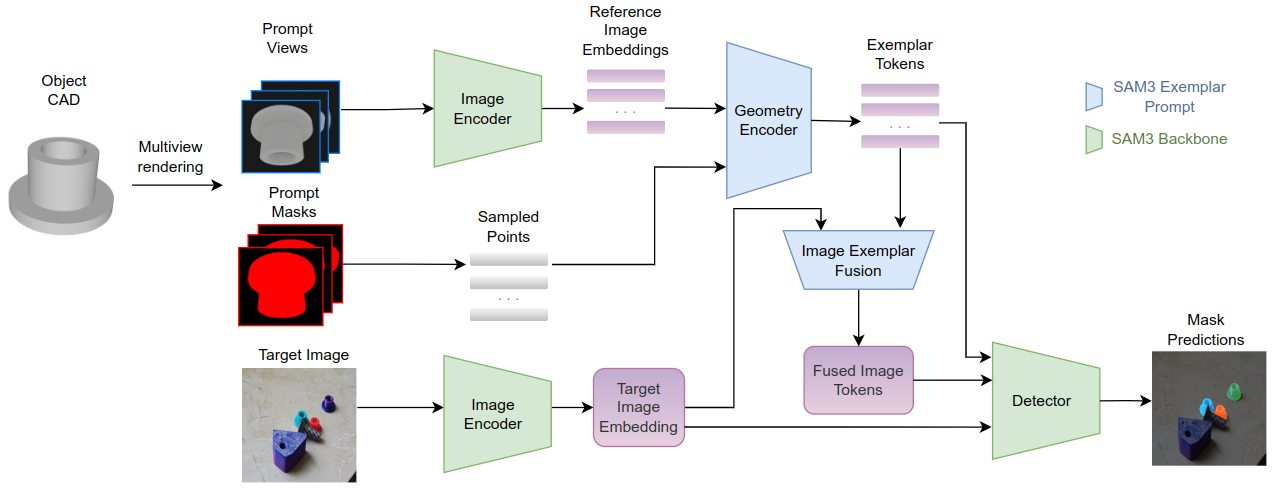}
    \caption{Model architecture. Canonical multi-view renderings of a CAD mesh are encoded into geometry-conditioned exemplar embeddings and concatenated as prompt tokens for SAM3's fusion transformer. The fused features are processed by the detector to produce instance masks in a single forward pass.}
    \label{fig:overview-figure}
\end{figure*}

We extend the SAM3 prompting framework to enable geometry-conditioned segmentation driven solely by CAD models. We first define our CAD-prompted segmentation task, then revisit the relevant components of SAM3, and finally introduce our geometry-based prompting mechanism and training procedure.

\subsection{Task Definition: CAD-Prompted Segmentation}

We consider the problem of segmenting all visible instances of a target object in a query RGB image given only its 3D mesh as specification.

Formally, given:
\begin{itemize}
    \item A target mesh $\mathcal{M}$,
    \item A query RGB image $I \in \mathbb{R}^{H \times W \times 3}$,
\end{itemize}
the goal is to predict binary masks $\{\hat{S}_k\}$ corresponding to all visible instances of $\mathcal{M}$ in $I$.

No textual description, category label, or real image exemplar is provided at inference time. The only object specification is the 3D geometry $\mathcal{M}$. The objective is to condition dense mask prediction directly on geometric specification within a single-stage, promptable segmentation framework.
\subsection{Prior Work: Revisiting SAM3} %prior work

SAM3 is a promptable segmentation model that predicts instance masks conditioned on input prompts. It consists of four primary components:

\begin{itemize}
    \item \textbf{Image Encoder:} extracts dense visual tokens from an input image.
    \item \textbf{Prompt Encoder:} converts prompts into embedding tokens, including \texttt{GeometryEncoder} for point prompts and \texttt{TextEncoder} for text prompts.
    \item \textbf{ImageExemplarFusion:} conditions image features on prompt embeddings via cross-attention.
    \item \textbf{Detector:} composed of a detection head and mask decoder, which predict instance tokens and decode them into pixel-level segmentation masks.
\end{itemize}

Given an image $I$, the \textbf{Image Encoder} extracts dense visual tokens. Prompt embeddings are then fused with the image features through the \texttt{ImageExemplarFusion} module, producing prompt-conditioned representations that are processed by the detector to generate instance masks.

SAM3 also supports exemplar prompting via the \texttt{GeometryEncoder}, which encodes point-conditioned prompt features and conditions query image features through the fusion transformer. In the official SAM3 formulation, exemplar prompting uses prompts originating from the same image. In our setting, we instead construct exemplar prompts from rendered views of a CAD model.

Our contribution lies in extending the prompt interface to support geometry-conditioned embeddings derived from CAD models.
\subsection{Geometry-Conditioned Prompt Construction}

\begin{figure}[t]
    \centering
    
    \begin{subfigure}{0.48\columnwidth}
        \centering
        \includegraphics[width=\columnwidth]{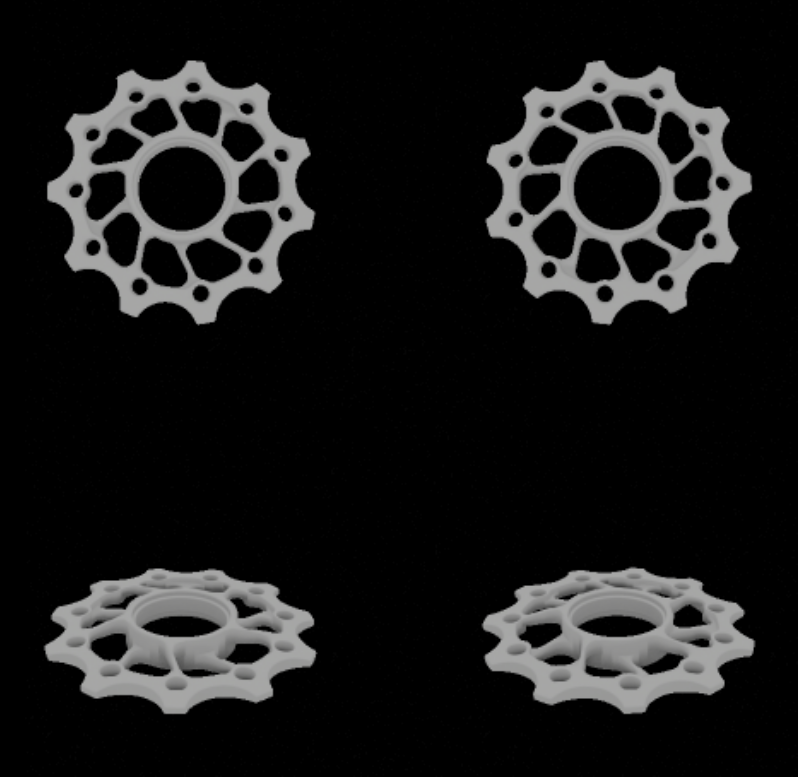}
        \caption{Multi-view CAD renderings as prompt.}
        \label{fig:multiview_prompt}
    \end{subfigure}
    \hfill
    \begin{subfigure}{0.48\columnwidth}
        \centering
        \includegraphics[width=\columnwidth]{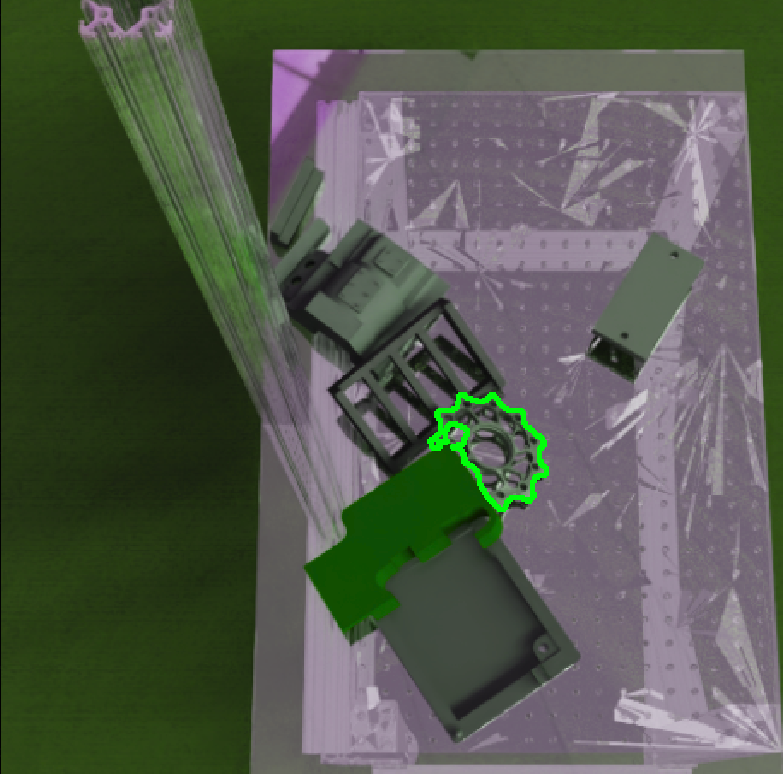}
        \caption{Synthetic training scene with GT contour.}
        \label{fig:synthetic_scene}
    \end{subfigure}
    
    \caption{Geometry-conditioned training pipeline. 
    (a) Canonical CAD renderings serve as prompts. 
    (b) Synthetic training data with instance annotations.}
    \label{fig:synthetic_pipeline}
\end{figure}

\paragraph{Multi-View Mesh Rendering}
%meeting_notes: explain 12 and blender 
Given a CAD mesh $\mathcal{M}$, we render it from predefined canonical viewpoints $V=12$ using Blender, although the method is not tied to a specific renderer. The viewpoints are distributed over the viewing sphere to capture diverse geometric perspectives. We empirically found 12 viewpoints sufficient to expose the major recognizable geometric features of most parts while maintaining manageable computational cost. Multi-view projections have been shown to provide an effective representation of 3D shape by encoding complementary geometric information across viewpoints \cite{mvcnn}. Subsequent work demonstrated that modeling relationships across views further enhances geometric encoding quality \cite{wei2019viewgcn}. Each rendering produces an RGB image $R_v$ and its corresponding foreground mask (Fig.~\ref{fig:synthetic_pipeline}a).

\paragraph{View Encoding via Geometry Encoder}

Each rendered view $R_v$ is encoded using the SAM3 image encoder:
\begin{align}
T_v = \text{ImageEncoder}(R_v).
\end{align}

We uniformly sample 25 point prompts within the foreground mask of each rendering. The image tokens $T_v$ and the point prompts $P_v$ are passed through the \texttt{SAM3 GeometryEncoder}:
\begin{align}
E_v = \text{GeometryEncoder}(T_v, P_v).
\end{align}

This yields a set of geometry-derived exemplar embeddings $\{E_v\}_{v=1}^{V}$ representing canonical views of the mesh.

\paragraph{Multi-View CAD Prompt Fusion}

In our setting, the exemplar prompts are derived from rendered views of a CAD model rather than directly from the query image. Given query image tokens
\begin{align}
T_I = \text{ImageEncoder}(I),
\end{align}
we compute
\begin{align}
T_{\text{fused}} = \text{ImageExemplarFusion}(T_I, \{E_v\}_{v=1}^{V}),
\end{align}
where $\{E_v\}_{v=1}^{V}$ are geometry-derived exemplar embeddings from $V=12$ rendered viewpoints.

The transformer-based fusion module naturally supports a variable number of prompt tokens, allowing multiple viewpoint embeddings to be concatenated without architectural modification. Self-attention within the transformer enables interaction among viewpoint tokens, allowing geometric information from different rendered views to be aggregated before conditioning the query image features.

Unlike conventional appearance exemplars, CAD renderings provide a geometry-centered representation of the target object that is less dependent on texture, lighting, and material appearance. This enables segmentation conditioned directly on geometric specification.
\subsection{Single-Stage Detection and Segmentation}

The fused image tokens $T_{\text{fused}}$ is processed by the inherited SAM3 detection head to generate instance tokens. Each instance token $D_k$ is decoded by the mask decoder to produce a segmentation mask and confidence score. The confidence score $s_k$ reflects the predicted confidence of the corresponding instance mask and can be used for prediction thresholding.
\begin{align}\label{eq:mask_decoder}
(\hat{S}_k, s_k) = \text{MaskDecoder}(T_{\text{fused}}, D_k).
\end{align}
%meeting_notes: confidence score explain and unify name
Detection and segmentation are performed in a single forward pass conditioned directly on the CAD-derived prompt.
\subsection{Training with Synthetic Data}

Although the backbone architecture is reused from SAM3, fine-tuning is required to align geometry-derived prompt embeddings with image features and to encourage geometry-focused reasoning. 

\paragraph{Synthetic Dataset Generation}
We collect approximately 9,000 CAD models from the ABC Dataset~\cite{abc}. Multi-object scenes are generated with Isaac Sim's Replicator pipeline by placing meshes in cluttered environments under diverse lighting conditions and background textures (Fig.~\ref{fig:synthetic_pipeline}b). The pipeline employs Automated Domain Randomization (ADR) across 21 independently configurable parameters spanning PBR material assignment, HDR lighting profiles, camera intrinsics and extrinsics, object scale and pose variation, and distractor placement. This randomization is designed to simulate the diversity of real manufacturing environments, including varying surface finishes, lighting conditions, and scene clutter, to prevent the model from exploiting appearance shortcuts and encourage geometry-based reasoning from CAD priors. Pixel-level ground-truth masks are obtained automatically from simulation. We generate approximately 15,000 training images, each containing 2 to 8 object instances. To encourage learning of instance-level separation, we design the sampling process such that half of the images contain at least two instances of the same CAD model, since uniform sampling from a large model pool would otherwise produce mostly single-instance cases.

\paragraph{Appearance Randomization}

To reduce reliance on color and texture cues, object materials, colors, and surface properties are randomly assigned independently of mesh identity. This encourages invariance with manufacturing variations and promotes geometry-driven segmentation.

\paragraph{Optimization Objective}
%meeting_notes: cite detail, add one sentence of what we do
Training is conducted in two stages to balance stability and multi-instance detection capability. Although DETR-style models often employ one-to-one matching to enforce distinct instance predictions and avoid post-processing~\cite{detr}, we found that strict one-to-one supervision leads to unstable optimization. We therefore adopt a two-stage training strategy that improves optimization stability while preserving multi-instance prediction capability.

\textbf{Stage 1: Score-Weighted Mask Supervision}
%reference back
In the initial stage, we supervise all predicted masks using a score-weighted mask loss. Each predicted instance from Eq.~\ref{eq:mask_decoder} consists of a mask $\hat{S}_k$ and a confidence score $s_k$. For each prediction, we compute a mask loss consisting of binary cross-entropy (BCE) and Dice loss:
\begin{align}
\mathcal{L}_{\text{mask}}^{(k)} = \mathcal{L}_{\text{BCE}}(\hat{S}_k, S^*_k) + \mathcal{L}_{\text{Dice}}(\hat{S}_k, S^*_k),
\end{align}
where $S^{*}_k$ denotes the ground-truth mask that best matches the prediction $\hat{S}_k$, defined as
\begin{align}
S^{*}_k = \arg\max_{S_j \in \mathcal{G}} \operatorname{IoU}(\hat{S}_k, S_j),
\end{align}
where $\mathcal{G}$ is the set of ground-truth instance masks in the image.

We weight each mask loss by its predicted confidence score:
\begin{align}
\mathcal{L}_{\text{stage1}} = \sum_k s_k \cdot \mathcal{L}_{\text{mask}}^{(k)}.
\end{align}

This formulation supervises all predicted outputs and provides strong gradient signals during early training, leading to improved optimization stability. However, since each prediction is independently encouraged to match any ground-truth mask without explicit instance-level assignment, this objective does not enforce coverage across distinct object instances. As a result, multiple predictions may collapse onto the same instance, while ignoring others when multiple instances are present.

\textbf{Stage 2: One-to-Many Matching.}

After the model learns basic geometric alignment between CAD prompts and image features, we switch to a one-to-many matching strategy to enable multi-instance prediction.

For each ground-truth instance, we allow up to $K=5$ predictions to be matched based on IoU-based greedy matching. The loss consists of matched mask loss (BCE + Dice) for selected predictions and presence loss supervising confidence scores:
\begin{align}
\mathcal{L}_{\text{stage2}} =
\sum_{k \in \mathcal{A}} \mathcal{L}_{\text{mask}}^{(k)} 
+ \mathcal{L}_{\text{presence}},
\end{align}
where $\mathcal{A}$ denotes the set of matched predictions. %meeting_notes M overloaded
The presence loss supervises confidence scores such that matched predictions are assigned a target label of 1, and unmatched predictions are assigned 0. Formally, for each prediction $k$, we define a binary target
\begin{align}
y_k =
\begin{cases}
1, & k \in \mathcal{A}, \\
0, & \text{otherwise},
\end{cases}
\end{align}
and compute a binary cross-entropy loss on the confidence score $s_k$:
\begin{align}
\mathcal{L}_{\text{presence}} =
\sum_{k} \mathcal{L}_{\text{BCE}}(s_k, y_k).
\end{align}
%cite NMS
This objective encourages the model to cover multiple object instances rather than collapsing predictions onto a single instance. During inference, non-maximum suppression (NMS) is applied to remove redundant overlapping predictions.

\subsection{Inference}

At inference time, only the target mesh $\mathcal{A}$ and the query image $I$ are required. The mesh is rendered into canonical views, encoded into geometry-conditioned prompt embeddings, fused with image features, and decoded in a single forward pass to produce segmentation masks.

This formulation extends SAM3 from language-conditioned and same-image exemplar prompting to cross-image, CAD-conditioned segmentation while preserving its single-stage, promptable architecture.

\section{Experimental Results}

\subsection{Experiment Setup}

\paragraph{Evaluation Protocol}
For all datasets, we adopt a prompt-based evaluation protocol. 
For each image, we issue one prompt per ground-truth object category present in that image. 
Each prompt is expected to segment all visible instances of the corresponding object within that image. 
This setup evaluates prompt-conditioned instance segmentation rather than closed-set detection over a predefined set of objects. 

To enable a fair comparison under this protocol, we make adaptations to baseline methods designed primarily for semantic segmentation or single-view prompting. For PerSAM, we aggregate multi-view CAD renderings using mean pooling to produce a single prompt embedding without modifying its architecture. For Matcher, we enable multiple exemplars and adapt its mask scoring to better suit instance segmentation, as its original formulation is tailored for semantic segmentation. These adaptations better align the baselines with the multi-view prompted instance segmentation setting, alongside results from their original formulations.

\paragraph{Metrics}
We report Panoptic Quality (PQ) and instance-level F1 score. 
PQ jointly measures segmentation accuracy and instance recognition quality, while F1 evaluates instance-level detection performance based on mask overlap. 
Together, these metrics capture both dense mask precision and multi-instance prediction accuracy.

In addition to instance-level evaluation, we also report semantic segmentation performance (mIoU) for comparison with prior exemplar-conditioned segmentation methods such as PerSAM and Matcher, which are primarily designed for semantic segmentation and do not distinguish between object instances. To obtain semantic predictions with CAD-Prompted SAM3, we merge all predicted instances with confidence scores above a fixed threshold corresponding to the query into a single binary mask. This conversion does not introduce additional learning or post-processing.
\begin{figure}
    \vspace{0.05in}
    \centering
    \includegraphics[width=0.4\textwidth] {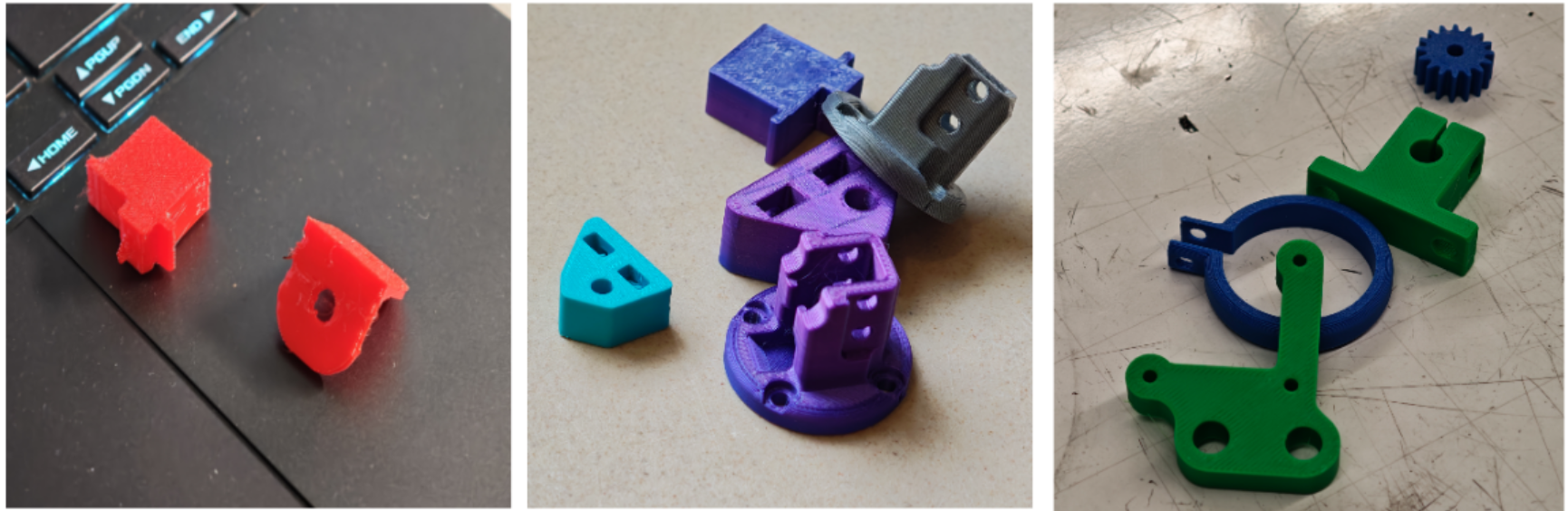}
    \caption{Representative samples from the custom 3D printing dataset.}
    \label{fig:3dprint_examples}
\end{figure}
\subsection{3D Printing Dataset}

To evaluate geometry-conditioned segmentation in rapid prototyping scenarios, we construct a custom dataset consisting of 20 distinct 3D-printed objects derived from CAD models. 
The dataset contains 200 real-world RGB images with an average of 4 object instances per image. Ground-truth instance masks are manually annotated for all objects.
Each object is printed in two different colors, and images are captured under diverse backgrounds with unrelated distractor objects (Fig.~\ref{fig:3dprint_examples}).

The objects do not appear in the synthetic data used for training the model, ensuring evaluation on unseen geometries.

Table~\ref{tab:3dprint_results} reports performance. 
Our method achieves the highest segmentation quality, obtaining a PQ of \textbf{0.754} and an F1 score of \textbf{0.803}. 
The strongest baseline, adapted Matcher, achieves a PQ of 0.514 and an F1 score of 0.573. 
PerSAM achieves a PQ of 0.318 and an F1 score of 0.341, while SAM3 exemplar prompting performs worse with a PQ of 0.141 and an F1 score of 0.156. Our method also achieves the highest mIoU, indicating strong semantic segmentation performance despite being optimized for instance segmentation.
These results demonstrate that geometry-conditioned prompting provides a significant advantage in rapid prototyping settings where object identity is defined primarily by shape rather than surface appearance.

\begin{table}[t]
\vspace{0.05in}
\centering
\caption{Performance on the custom 3D printing dataset.}
\label{tab:3dprint_results}
\begin{tabular*}{\linewidth}{@{\extracolsep{\fill}} lccc}
\hline
Method & PQ$\uparrow$ & F1$\uparrow$ & mIoU$\uparrow$ \\
\hline

PerSAM (Original) & 0.314 & 0.337 & 0.250 \\
PerSAM (Adapted) & 0.318 & 0.341 & 0.240 \\
\hline

Matcher (Original) & 0.249 & 0.280 & 0.443 \\
Matcher (Adapted) & 0.514 & 0.573 & 0.446 \\
\hline

SAM3 Exemplar Prompt & 0.141 & 0.156 & 0.132 \\
\hline

\textbf{CAD-Prompted SAM3 (Ours)} & \textbf{0.754} & \textbf{0.803} & \textbf{0.749} \\

\hline
\end{tabular*}
\end{table}
%meeting_notes: add semnatic evaluation picture
\subsection{Benchmark Datasets}

We further evaluate on two standard industrial benchmarks: T-LESS~\cite{hodan2017tless} and ITODD~\cite{drost2017itodd}. All benchmark results are reported on the official test splits.

\paragraph{T-LESS.}
T-LESS is an industry-related object dataset consisting of texture-less rigid components. 
The objects have no identifiable color or texture, many exhibit strong geometric similarity, and scenes are highly cluttered with significant occlusion. 
These characteristics make instance discrimination challenging for appearance-based methods.

\paragraph{ITODD.}
ITODD is an industrial setting dataset featuring texture-less metallic objects captured in grayscale. 
The dataset includes both uncluttered and highly cluttered scenes, emphasizing shape-based reasoning under limited appearance cues.

Table~\ref{tab:benchmark_results} summarizes benchmark performance. 
On T-LESS, our method achieves the highest PQ (0.317) and F1 (0.371), outperforming all baselines. 
On ITODD, our approach yields a substantial improvement, reaching PQ 0.536 and F1 0.657. Our approach also consistently achieves the highest mIoU across both benchmarks.
The consistent gains across both datasets indicate that conditioning segmentation directly on canonical CAD geometry improves robustness in industrial object perception.

\begin{table}[t]
\vspace{0.05in}
\centering
\setlength{\tabcolsep}{3.1pt}
\caption{Performance on T-LESS and ITODD benchmarks.}
\label{tab:benchmark_results}
\begin{tabular*}{\linewidth}{@{\extracolsep{\fill}} p{2.8cm}cccccc}
\hline
Method & \multicolumn{3}{c}{T-LESS} & \multicolumn{3}{c}{ITODD} \\
& PQ$\uparrow$ & F1$\uparrow$ & mIoU$\uparrow$ & PQ$\uparrow$ & F1$\uparrow$ & mIoU$\uparrow$ \\
\hline

PerSAM (Original) & 0.168 & 0.190 & 0.080 & 0.347 & 0.419 & 0.322 \\
PerSAM (Adapted) & 0.166 & 0.188 & 0.082 & 0.363 & 0.438 & 0.322 \\
\hline

Matcher (Original) & 0.163 & 0.184 & 0.248 & 0.263 & 0.357 & 0.631 \\
Matcher (Adapted) & 0.173 & 0.219 & 0.177 & 0.277 & 0.379 & 0.478 \\
\hline

SAM3 Exemplar Prompt & 0.171 & 0.189 & 0.138 & 0.415 & 0.507 & 0.415 \\
\hline

\textbf{CAD-Prompted SAM3 (Ours)} & \textbf{0.317} & \textbf{0.371} & \textbf{0.327} & \textbf{0.536} & \textbf{0.657} & \textbf{0.702} \\
\hline

\end{tabular*}
\end{table}
\begin{figure}
    \centering
    \includegraphics[width=0.50\textwidth] {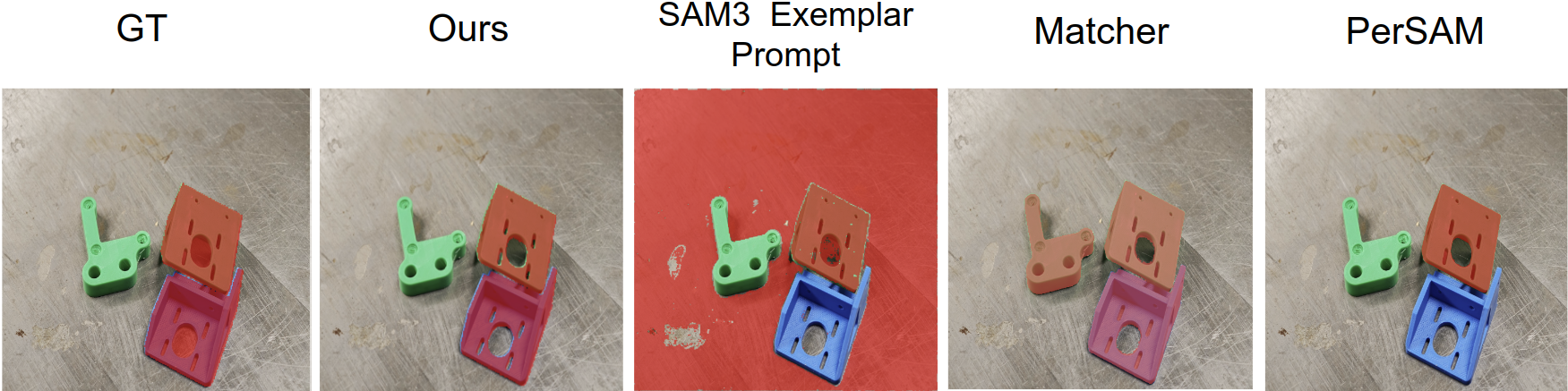}
    \caption{Qualitative comparision of CAD-Prompted SAM3 against baselines in semantic segmentation task}
    \label{fig:semantic_results}
\end{figure}
\begin{table}[b]
\centering
\begin{tabular}{l c c}
\hline
\textbf{Object} & \textbf{Ours} & \textbf{COPA} \\
\hline
Shaft Mount   & 9/10 & 1/10 \\
Gear          & 8/10 & 2/10 \\
Square Tube   & 8/10 & 3/10 \\
Bolt          & 6/10 & 2/10 \\
LEGO Brick    & 9/10 & 3/10 \\
\hline
\end{tabular}
\caption{Pick-and-place success rates for different objects, comparing our method with COPA~\cite{copa2024}.}
\label{tab:pick_place_results}
\end{table}
\begin{figure*} [t]
\vspace{0.05in}
    \centering

    \begin{subfigure}[t]{0.18\textwidth}
        \includegraphics[width=\linewidth]{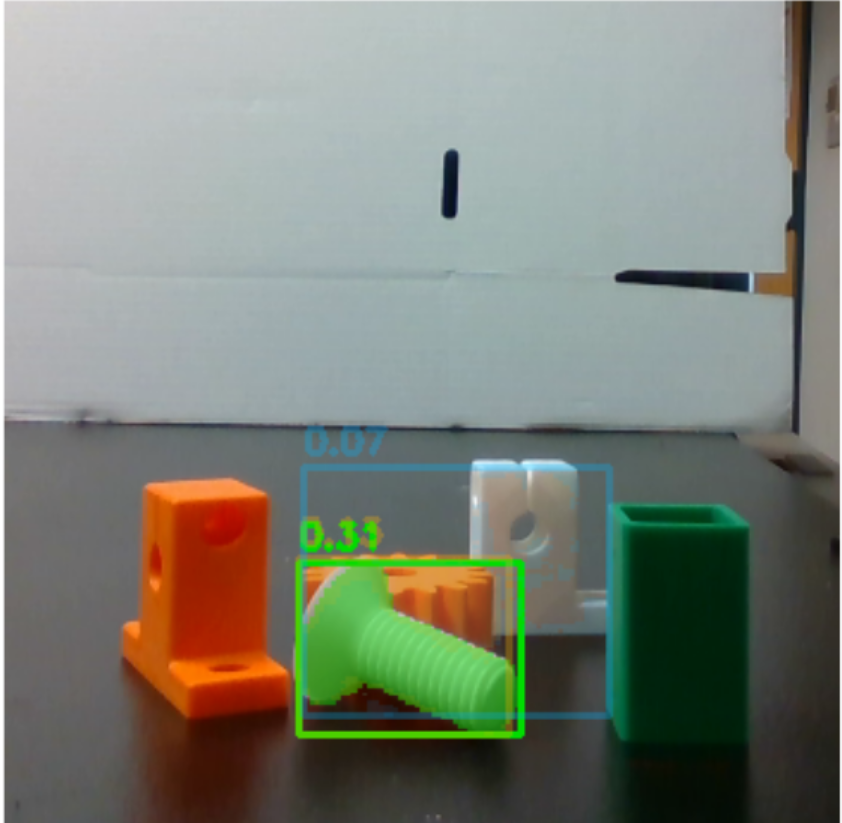}
        \caption{Instance Segmentation}
        \label{fig:compute_offset_s1}
    \end{subfigure}\hfill
    \begin{subfigure}[t]{0.18\textwidth}
        \includegraphics[width=\linewidth]{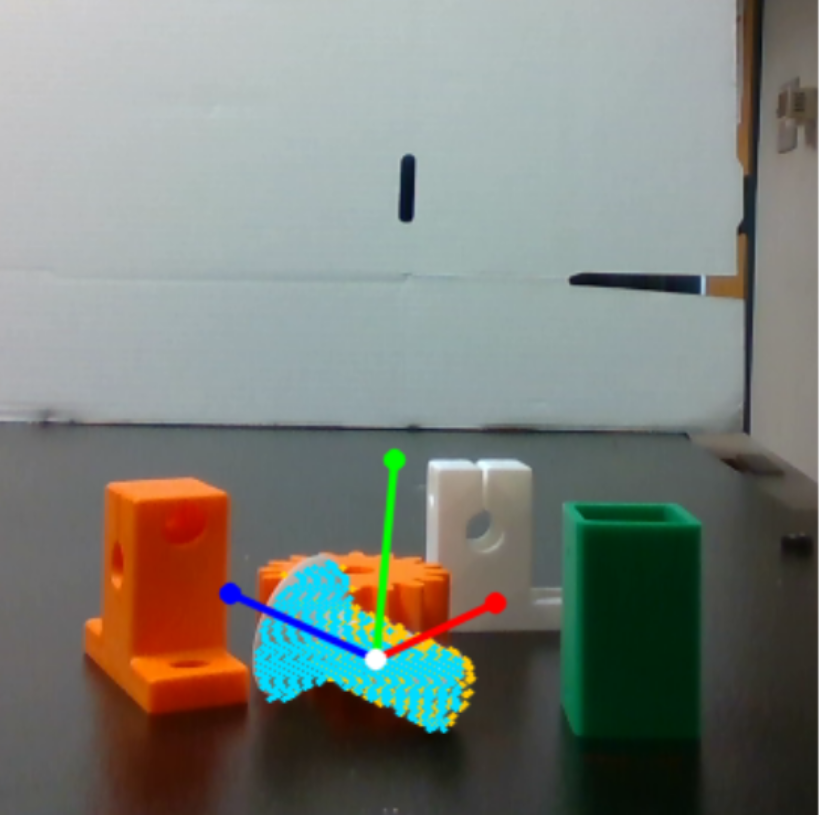}
        \caption{Pose Estimation}
        \label{fig:compute_offset_s2}
    \end{subfigure}\hfill
    \begin{subfigure}[t]{0.18\textwidth}
        \includegraphics[width=\linewidth]{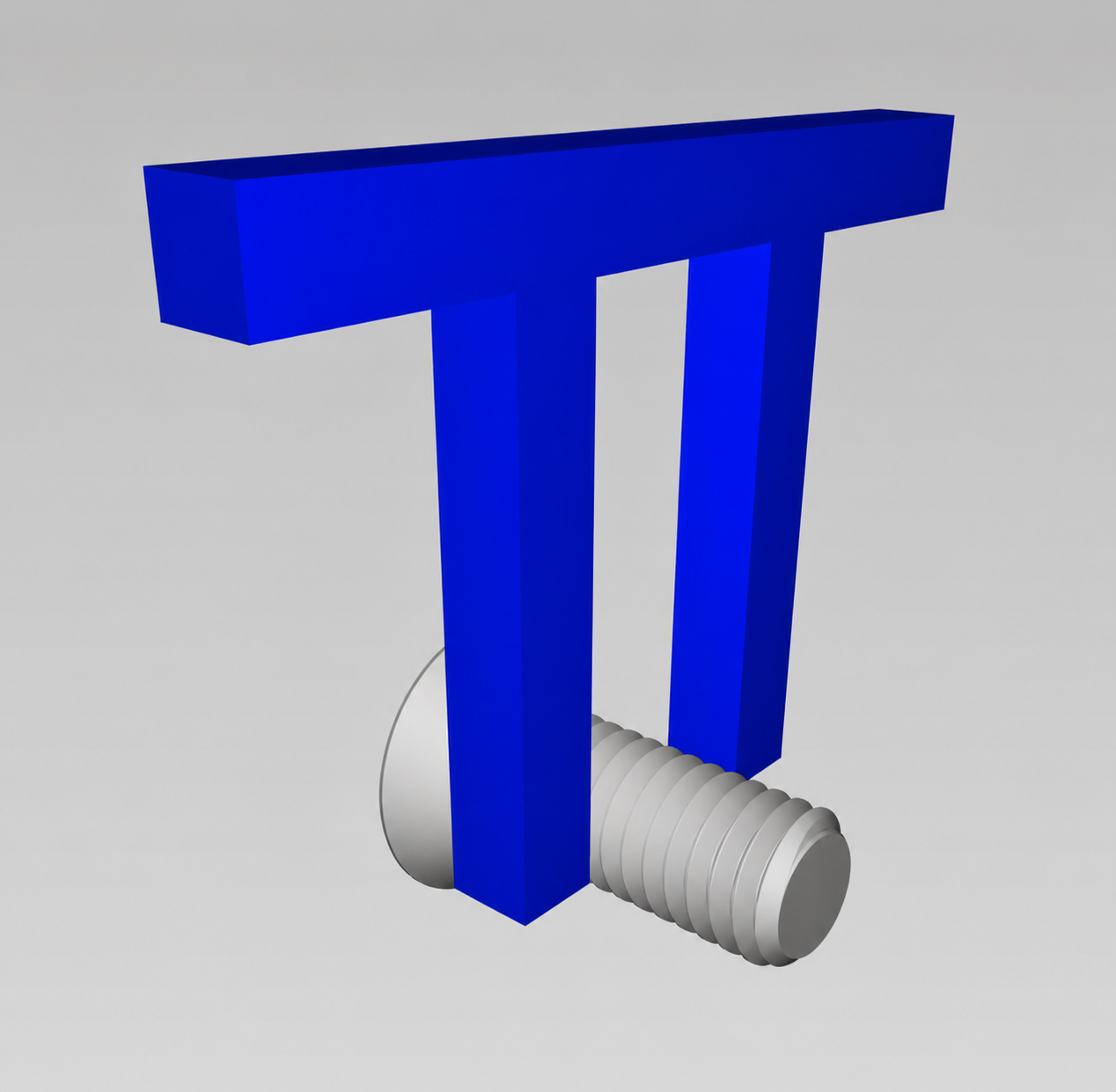}
        \caption{Grasp Generation}
        \label{fig:compute_offset_s3}
    \end{subfigure}\hfill
    \begin{subfigure}[t]{0.41\textwidth}
        \includegraphics[width=\linewidth]{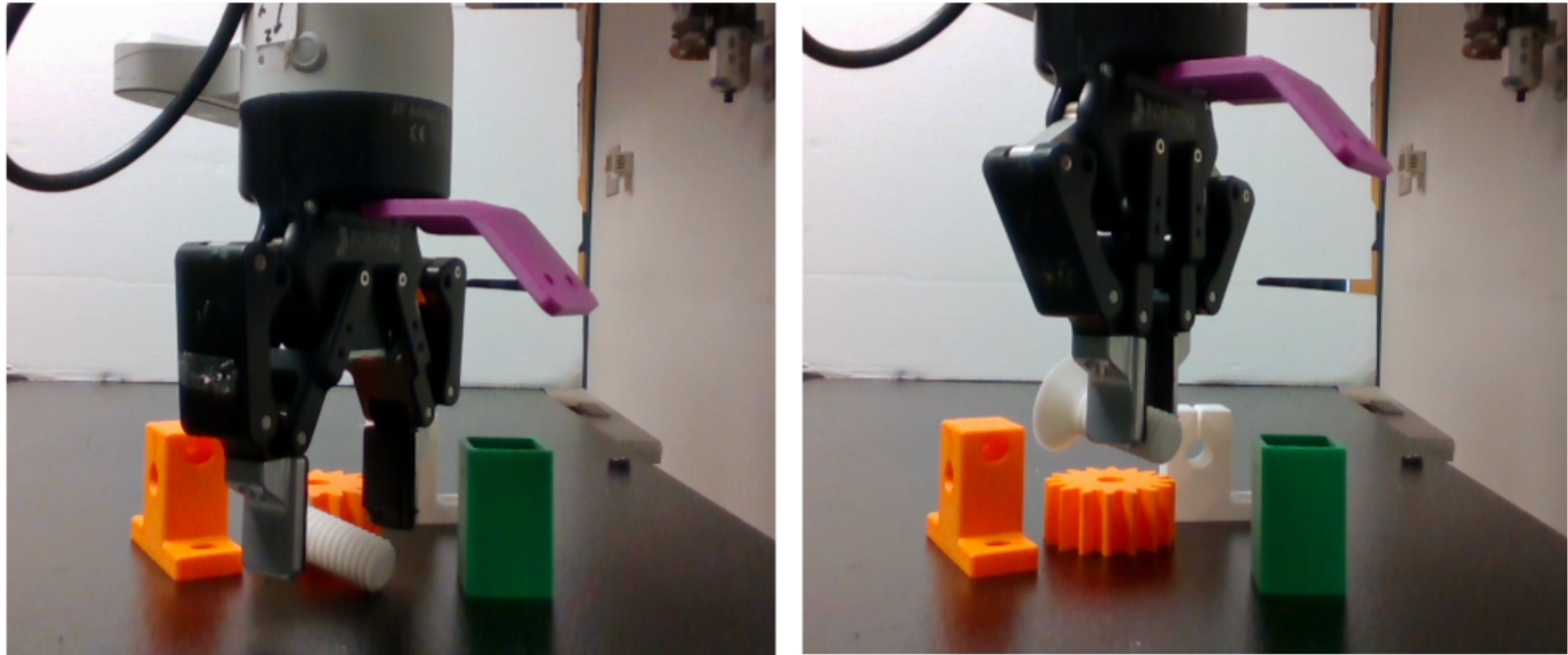}
        \caption{Robot Execute Grasp}
        \label{fig:compute_offset_s4}
    \end{subfigure}\hfill

    \caption{Zero-shot Manipulation Pipeline based on CAD-Prompted SAM3}
    \label{fig:manipulation_pipeline}
    \vspace{-15pt}
\end{figure*}
\subsection{Applying to Zero-Shot Manipulation}

Beyond instance segmentation, the predicted masks can be directly used as geometric filters for downstream tasks such as pose estimation and grasp planning. 
To illustrate this, we demonstrate a simple zero-shot manipulation (Fig.~\ref{fig:manipulation_pipeline}) built on top of our method.

Given an RGB-D observation and a set of CAD models, the system performs:

(1) Instance segmentation using CAD-prompted SAM3.

(2) Pose estimation via ICP on segmented point clouds \cite{besl1992method}.

(3) Grasp generation from the CAD models, transformed by the estimated pose, using a simple physics-based antipodal grasp formulation \cite{murray1994mathematical}.

All stages rely solely on CAD models, requiring no object-specific training or labeled real images.

\paragraph{Setup.}
Experiments are conducted using a fixed RGB-D camera and a 6-DoF robot arm with a parallel-jaw gripper. 
Objects are placed in moderately cluttered tabletop scenes, and the robot is tasked with sorting objects into designated bins. 
Only CAD models are available for all objects; no labeled real images or object-specific annotations or parameters are used.
We compare against COPA~\cite{copa2024}, a zero-shot manipulation pipeline that first uses SAM~\cite{sam1} to generate candidate segmentation masks for all potential objects, and then selects the target via vision-language prompting. It predicts grasps directly from raw point clouds using GraspNet~\cite{graspnet}, and selects those that lie on the chosen object. 
At test time, our method is conditioned on CAD models, while COPA is conditioned on textual descriptions.

\paragraph{Results.}
Table~\ref{tab:pick_place_results} reports pick-and-place success rates. 
Our method achieves consistently higher success across all objects, with an average success rate of approximately 80\%, compared to 22\% for COPA. 
In addition, our pipeline is approximately 15$\times$ faster than COPA, as it avoids exhaustive segmentation and reliance on language model API calls. 
These results demonstrate that geometry-conditioned segmentation provides a strong and efficient basis for zero-shot manipulation.

\begin{table}[t]
\setlength{\tabcolsep}{3pt} % default is ~6pt
\centering
\begin{tabular}{l l c l c}
\hline
\multirow{2}{*}{\textbf{Stage}} & \multicolumn{2}{c}{\textbf{Ours}} & \multicolumn{2}{c}{\textbf{COPA}} \\
\cline{2-5}
 & \textbf{Module} & \textbf{Time (s)} & \textbf{Module} & \textbf{Time (s)} \\
\hline
\multirow{2}{*}{Detection}
& \multirow{2}{*}{\begin{tabular}{@{}l@{}} CAD-Prompted \\ SAM3 \end{tabular}} 
& \multirow{2}{*}{0.170}
& SAM & 5.80 \\
&  &  & GPT5.4 API & 1.29 \\
\hline
Pose Estimation
& ICP             & 0.112 & --               & -- \\
\hline
Grasp Prediction
& Physics-Based   & 0.188 & GraspNet         & 0.172 \\
\hline
\textbf{Total}
& \multicolumn{2}{c}{\textbf{0.460}} & \multicolumn{2}{c}{\textbf{7.26}} \\
\hline
\end{tabular}
\caption{Per-object average runtime breakdown for our method and COPA. Both methods are evaluated on a computer with NVIDIA RTX 4070 GPU and Intel Core Ultra 9 185H CPU.}
\label{tab:runtime_comparison}
\end{table}

\section{Conclusion and Future Work}

\subsection{Conclusion}

We presented a geometry-conditioned promptable segmentation framework that extends foundation segmentation models to industrial objects defined by CAD specifications. 
Instead of relying on language descriptions or RGB exemplars, our method conditions segmentation directly on canonical multi-view renderings derived from CAD models, enabling shape-based object identification independent of surface appearance.

Built upon SAM3, our approach formulates CAD-conditioned perception as single-stage promptable instance segmentation. 
Across a custom 3D printing dataset and standard industrial benchmarks including T-LESS and ITODD, CAD-Prompted SAM3 consistently improves segmentation quality over appearance-based exemplar methods. 
The gains are particularly pronounced in rapid prototyping scenarios where objects vary in color and context but retain fixed geometry.

These results demonstrate that canonical geometry can serve as a robust and expressive prompt modality for foundation segmentation models. 
By aligning perception with the way industrial objects are defined, i.e., through precise CAD models rather than textual labels or visual exemplars, our formulation offers a practical solution for perception for rapid prototyping and small-batch manufacturing.

\subsection{Future Work: Cross-Modal Geometry Conditioning}

While this work uses multi-view renderings of CAD models as prompts, an important next step is to move beyond rasterized images and explore direct geometric conditioning. 
Future work could investigate encoding meshes, point clouds, or implicit surface representations directly into the prompt space of foundation models, enabling tighter integration between geometric reasoning and perception, without the middle step of canonical rendering.

\bibliographystyle{IEEEtran}
\bibliography{reference}

@article{sam3,
  title        = {SAM 3: Segment Anything with Concepts},
  author       = {Carion, Nicolas and Gustafson, Laura and Hu, Yuan‐Ting and Debnath, Shoubhik and Hu, Ronghang and Suris, Didac and Ryali, Chaitanya and Vasudev Alwala, Kalyan and Khedr, Haitham and Huang, Andrew and Lei, Jie and Ma, Tengyu and Guo, Baishan and Kalla, Arpit and Marks, Markus and Greer, Joseph and Wang, Meng and Sun, Peize and R{\"a}dle, Roman and Afouras, Triantafyllos and Mavroudi, Effrosyni and Xu, Katherine and Wu, Tsung‐Han and Zhou, Yu and Momeni, Liliane and Hazra, Rishi and Ding, Shuangrui and Vaze, Sagar and Porcher, Francois and Li, Feng and Li, Siyuan and Kamath, Aishwarya and Cheng, Ho Kei and Doll{\'a}r, Piotr and Ravi, Nikhila and Saenko, Kate and Zhang, Pengchuan and Feichtenhofer, Christoph},
  journal      = {arXiv preprint},
  volume       = {arXiv:2511.16719},
  year         = {2025},
  url          = {https://arxiv.org/abs/2511.16719},
}

@inproceedings{matcher,
  title        = {Matcher: Segment Anything with One Shot Using All‐Purpose Feature Matching},
  author       = {Liu, Yang and Zhu, Muzhi and Li, Hengtao and Chen, Hao and Wang, Xinlong and Shen, Chunhua},
  booktitle    = {International Conference on Learning Representations (ICLR) 2024},
  year         = {2024},
  url          = {https://proceedings.iclr.cc/paper_files/paper/2024/hash/4df9a5e6bad9e64ebcea453e031142bb-Abstract-Conference.html},
}

@inproceedings{cartman,
  title={Cartman: The Low-Cost Cartesian Manipulator that Won the Amazon Robotics Challenge},
  author={Morrison, Dim and Tow, Adam W. and McTaggart, Matthew and Smith, R. and Kelly-Boxall, N. and Wade-McCue, S. and Erskine, J. and Grinover, R. and Gurman, A. and Hunn, T. and Lee, D. and Milan, A. and Pham, T. and Rallos, G. and Razjigaev, A. and Rowntree, T. and Vijay, K. and Zhuang, Z. and Lehnert, C. and Reid, I. and Corke, P. and Leitner, J.},
  booktitle={Proceedings of the IEEE International Conference on Robotics and Automation (ICRA)},
  year={2018},
  doi={10.1109/ICRA.2018.8463191}
}

@inproceedings{mvcnn,
  title={Multi-View Convolutional Neural Networks for 3D Shape Recognition},
  author={Su, Hang and Maji, Subhransu and Kalogerakis, Evangelos and Learned-Miller, Erik},
  booktitle={Advances in Neural Information Processing Systems (NeurIPS)},
  year={2015}
}

@inproceedings{wei2019viewgcn,
  title={View-GCN: View-Based Graph Convolutional Network for 3D Shape Analysis},
  author={Wei, Xin and Shen, Chunhua and Wang, Yan and others},
  booktitle={Proceedings of the IEEE/CVF Conference on Computer Vision and Pattern Recognition (CVPR)},
  year={2019}
}

@inproceedings{abc,
  title={ABC: A Big CAD Model Dataset for Geometric Deep Learning},
  author={Koch, Sebastian and Matveev, Albert and Jiang, Zhongshi and Williams, Francis and Artemov, Alexey and Burnaev, Evgeny and Alexa, Marc and Zorin, Denis and Panozzo, Daniele},
  booktitle={Proceedings of the IEEE Conference on Computer Vision and Pattern Recognition (CVPR)},
  year={2019}
}

@inproceedings{hodan2017tless,
  title={{T-LESS}: An {RGB-D} Dataset for 6D Pose Estimation of Texture-less Objects},
  author={Hoda{\v{n}}, Tom{\'a}{\v{s}} and Haluza, Pavel and Obdr{\v{z}}{\'a}lek, {\v{S}}t{\v{e}}p{\'a}n and Matas, Ji{\v{r}}{\'\i} and Lourakis, Manolis and Zabulis, Xenophon},
  booktitle={2017 IEEE Winter Conference on Applications of Computer Vision (WACV)},
  pages={880--888},
  year={2017}
}

@inproceedings{drost2017itodd,
  title={Introducing {MVTec} {ITODD} — A Dataset for 3D Object Recognition in Industry},
  author={Drost, Bertram and Ulrich, Markus and Bergmann, Paul and H{\"a}rtinger, Philipp and Steger, Carsten},
  booktitle={2017 IEEE International Conference on Computer Vision Workshops (ICCVW)},
  pages={2200--2208},
  year={2017}
}

@article{sam1,
  title={Segment Anything},
  author={Kirillov, Alexander and Mintun, Eric and Ravi, Nikhila and others},
  journal={arXiv preprint arXiv:2304.02643},
  year={2023}
}

@inproceedings{tian2020pfenet,
  title={Prior Guided Feature Enrichment Network for Few-Shot Segmentation},
  author={Tian, Zhuotao and Zhao, Hengshuang and others},
  booktitle={CVPR},
  year={2020}
}

@article{zhang2023persam,
  title={Personalize Segment Anything Model with One Shot},
  author={Zhang, Renrui and others},
  journal={arXiv preprint arXiv:2305.03048},
  year={2023}
}

@inproceedings{cnos,
  title        = {CNOS: A Strong Baseline for CAD-Based Novel Object Segmentation},
  author       = {Nguyen, Thanh-Tuan and Lee, Jihoon and Cho, Myungin and Canny, John F.},
  booktitle    = {Proceedings of the IEEE/CVF International Conference on Computer Vision Workshops (ICCVW) R6D Workshop},
  year         = {2023},
  pages        = {338–350},
  publisher    = {IEEE/CVF},
  url          = {https://openaccess.thecvf.com/content/ICCV2023W/R6D/papers/Nguyen_CNOS_A_Strong_Baseline_for_CAD-Based_Novel_Object_Segmentation_ICCVW_2023_paper.pdf}
}

@inproceedings{canet,
  title={CANet: Class-agnostic Segmentation Networks with Iterative Refinement and Attentive Few-Shot Learning},
  author={Zhang, Chi and Lin, Guosheng and Liu, Fayao and Yao, Rui and Shen, Chunhua},
  booktitle={Proceedings of the IEEE/CVF Conference on Computer Vision and Pattern Recognition (CVPR)},
  pages={5217--5226},
  year={2019}
}

@article{dinov2,
  title={DINOv2: Learning Robust Visual Features without Supervision},
  author={Oquab, Maxime and Darcet, Timoth{\'e}e and Moutakanni, Th{\'e}o and Vo, Huy V. and Szafraniec, Marc and Khalidov, Vasil and Fernandez, Pierre and Haziza, Daniel and Massa, Francisco and El-Nouby, Alaaeldin and Assran, Mahmoud and Ballas, Nicolas and Galuba, Wojciech and Howes, Russell and Huang, Po-Yao and Li, Shang-Wen and Misra, Ishan and Rabbat, Michael and Sharma, Vasu and Synnaeve, Gabriel and Xu, Hu and J{\'e}gou, Herv{\'e} and Mairal, Julien and Labatut, Patrick and Joulin, Armand and Bojanowski, Piotr},
  journal={arXiv preprint arXiv:2304.07193},
  year={2023}
}

@inproceedings{dinov1,
  title={Emerging Properties in Self-Supervised Vision Transformers},
  author={Caron, Mathilde and Touvron, Hugo and Misra, Ishan and J{\'e}gou, Herv{\'e} and Mairal, Julien and Bojanowski, Piotr and Joulin, Armand},
  booktitle={Proceedings of the IEEE/CVF International Conference on Computer Vision (ICCV)},
  year={2021}
}

@article{nids-net,
  title        = {Neural Instance Detection and Segmentation Network (NIDS-Net)},
  author       = {Li, Xin and Zhang, Qiang and Wang, Yu and Chen, Kai},
  journal      = {arXiv preprint},
  volume       = {arXiv:2405.17859},
  year         = {2024},
  url          = {https://arxiv.org/abs/2405.17859}
}

@article{muse,
author = {Cho, Sungmin and Park, Sungbum and Oh, Insoo},
year = {2025},
month = {10},
pages = {},
title = {MUSE: Model-based Uncertainty-aware Similarity Estimation for zero-shot 2D Object Detection and Segmentation},
doi = {10.48550/arXiv.2510.17866}
}

@inproceedings{peng2024finegrained,
  title={Synthesize, Diagnose, and Optimize: Towards Fine-Grained Vision-Language Understanding},
  author={Peng, X. and others},
  booktitle={Proceedings of the IEEE/CVF Conference on Computer Vision and Pattern Recognition (CVPR)},
  year={2024}
}

@article{ju2025evaluation,
  title={Evaluation of Vision-Language Models under Fine-Grained Visual Discrimination Tasks},
  author={Ju, X. and others},
  journal={Applied Sciences},
  year={2025},
  volume={15},
  number={9}
}

@article{chen2024learning,
  title={Learning Self-Target Knowledge for Few-Shot Segmentation},
  author={Chen, Yadang and Chen, Sihan and Yang, Zhi-Xin and Wu, Enhua},
  journal={Pattern Recognition},
  volume={149},
  pages={110266},
  year={2024},
  publisher={Elsevier}
}

@incollection{robocup,
  author       = {Raphael Memmesheimer and Jan Nogga and Bastian P\"atzold and Evgenii Kruzhkov and Simon Bultmann and Michael Schreiber and Jonas Bode and Bertan Karacora and Juhui Park and Alena Savinykh and Sven Behnke},
  title        = {RoboCup@Home 2024 OPL Winner NimbRo: Anthropomorphic Service Robots Using Foundation Models for Perception and Planning},
  booktitle    = {Robot World Cup. RoboCup 2024},
  series       = {Lecture Notes in Computer Science},
  publisher    = {Springer},
  year         = {2025},
  pages        = {515--527},
  doi          = {10.1007/978-3-031-85859-8\_44}
}

@article{copa2024,
  title={CoPa: General Robotic Manipulation through Spatial Constraints of Parts with Foundation Models},
  author={Huang, Haoxu and Lin, Fanqi and Hu, Yingdong and Wang, Shengjie and Gao, Yang},
  journal={arXiv preprint arXiv:2403.08248},
  year={2024}
}

@book{murray1994mathematical,
  title={A Mathematical Introduction to Robotic Manipulation},
  author={Murray, Richard M and Li, Zexiang and Sastry, S Shankar},
  year={1994},
  publisher={CRC press}
}

@article{besl1992method,
  title={A method for registration of 3-D shapes},
  author={Besl, Paul J and McKay, Neil D},
  journal={IEEE TPAMI},
  year={1992}
}

@inproceedings{graspnet,
  title={GraspNet-1Billion: A Large-Scale Benchmark for General Object Grasping},
  author={Fang, Hao-Shu and Wang, Chen and Gou, Minghao and Lu, Cewu},
  booktitle={CVPR},
  year={2020}
}

@inproceedings{detr,
  author    = {Nicolas Carion and Francisco Massa and Gabriel Synnaeve and Nicolas Usunier and Alexander Kirillov and Sergey Zagoruyko},
  title     = {End-to-End Object Detection with Transformers},
  booktitle = {European Conference on Computer Vision (ECCV)},
  year      = {2020}
}

@article{yolo,
  author  = {Joseph Redmon and Santosh Divvala and Ross Girshick and Ali Farhadi},
  title   = {You Only Look Once: Unified, Real-Time Object Detection},
  journal = {Proceedings of the IEEE Conference on Computer Vision and Pattern Recognition (CVPR)},
  year    = {2016}
}

\end{document}